\documentclass{article}
\usepackage{amsmath,epsfig}
\usepackage{hyperref}
\usepackage[table,dvipsnames]{xcolor}

\usepackage[preprint]{spconf}
\copyrightnotice{\copyright\ IEEE 2018}
\toappear{To appear in {\it Proc.\ IGARSS 2018,
July 22-27, 2018, Valencia, Spain}}

\usepackage{xcolor}

\title{Urban Change Detection for Multispectral Earth Observation \\ Using Convolutional Neural Networks}
\name{Rodrigo Caye Daudt\textsuperscript{1,2}, Bertrand Le Saux\textsuperscript{1}, Alexandre Boulch\textsuperscript{1}, Yann Gousseau\textsuperscript{2}}
\address{\textsuperscript{1}DTIS, ONERA, Universit\'{e} Paris-Saclay, FR-91123 Palaiseau, France \\ \textsuperscript{2}LTCI, T\'{e}l\'{e}com ParisTech, FR-75013 Paris, France}
\begin{document}

\maketitle

\begin{abstract}

The Copernicus Sentinel-2 program now provides multispectral images at a global scale with a high revisit rate. In this paper we explore the usage of convolutional neural networks for urban change detection using such multispectral images. We first present the new change detection dataset that was used for training the proposed networks, which will be openly available to serve as a benchmark. The Onera Satellite Change Detection (OSCD) dataset is composed of pairs of multispectral aerial images, and the changes were manually annotated at pixel level. 
We then propose two architectures to detect changes, Siamese and Early Fusion, and compare the impact of using different numbers of spectral channels as inputs.
These architectures are trained from scratch using the provided dataset.

\end{abstract}

\section{Introduction}
\label{sec:intro}

Change detection is an integral part of the analysis of satellite imagery, and it has been studied for several decades \cite{hussain2013change,singh1989review}. It consists of comparing a registered pair of images of the same region and identifying the parts where a change has occurred, e.g. vegetation evolution or urban changes. A label is assigned to each pixel: change or no change. The nature of the changes that are detected may vary with the desired application, such as vegetation changes or urban changes (artificialization). Change detection is a crucial step for analysing temporal Earth observation sequences in order to build evolution maps of land cover, urban expansion, deforestation, etc.

With the rise of open access Earth observation from programs such as Copernicus and Landsat, large amounts of data are available. The Sentinel-2 satellites generate time series of multispectral images of Earth's landmasses with resolutions varying between 10m and 60m. Despite the abundance of raw data, there is a lack of open labelled datasets using these images which can be used for quantitative comparison and evaluation of new proposed change detection algorithms. Labelled datasets are also necessary for developing supervised learning methods, which have in the last few years been used to achieve the state-of-the-art results in many problems in the areas of computer vision and image processing. Machine learning techniques such as Convolutional Neural Networks (CNNs) have been on the rise not only due to the exponential growth of the available computing power, but also to the increasingly large amounts of available data. These techniques can not be adequately applied to the problem of change detection while there is a lack of data that can be used for training these systems.

\subsection{Related Work}
\label{ssec:rw}

Change detection has a long history. The first techniques that were proposed used manually crafted processes to identify changes, while later methods have proposed using machine learning algorithms in different ways \cite{hussain2013change,singh1989review,le2013urban,liu2016unsupervised}. Recent advances in machine learning algorithms for image analysis have not yet taken over the area of change detection due to the lack of large amounts of training data. Thus, some methods have been proposed recently to use transfer learning to circumvent this problem~\cite{el2017zoom}. While transfer learning is a valid option, it may limit the reach of the proposed methods. For example, the vast majority of the large CNNs trained on big datasets use RGB images, while the Sentinel-2 images contain 13 useful bands, most of which would need to be ignored when using such a system. Moreover, the recently proposed deep learning algorithms for change detection have mostly been designed to generate a difference image which is manually thresholded \cite{zhan2017change}. This avoids end-to-end training, which tends to achieve better results and faster execution. Related to change detection, deep learning techniques have been developed for computer vision applications with the aim of comparing image pairs \cite{zagoruyko2015learning}.

\subsection{Contributions}
\label{ssec:contrib}

The aim of this paper is twofold. The first contribution that is presented in this paper is the development of an urban change detection dataset of image pairs and pixel-wise labels to be used for training, testing and comparing change detection algorithms: the Onera Satellite Change Detection (OSCD) dataset will be openly available on the internet\footnote{\url{http://dase.grss-ieee.org/}}. This dataset was created using the multispectral images taken by the Sentinel-2 satellites of places with different levels of urbanization in several different countries that have experienced urban growth and/or changes. The second contribution presented in this paper is the proposal of two different CNN architectures that aim to learn end-to-end change detection from this dataset in a fully supervised manner.

This paper is structured as follows. Section~\ref{sec:dataset} describes the methods and challenges for creating the urban change detection dataset, as well as information about what is contained in it. Section~\ref{sec:cd} describes the proposed supervised learning methods used for change detection. Section~\ref{sec:results} presents the results for different tests which explore the reach and limitations of both the dataset and the proposed methods.

\section{Dataset}
\label{sec:dataset}

\begin{figure*}[ht]

  \begin{minipage}[b]{0.18\linewidth}
    \centering
    \centerline{\epsfig{figure=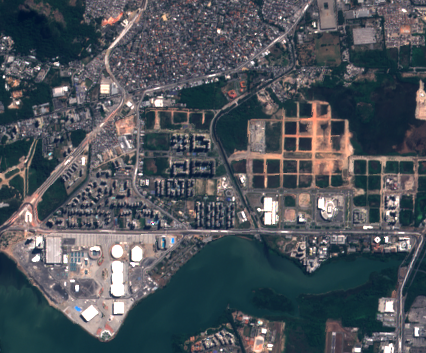,width=0.8\linewidth}}
    \centerline{(a) Rio in 24/04/2016.}\medskip
  \end{minipage}
  \hfill
  \begin{minipage}[b]{0.18\linewidth}
    \centering
    \centerline{\epsfig{figure=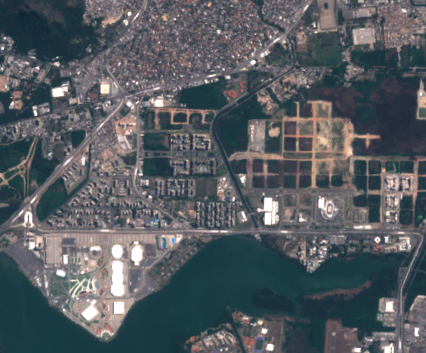,width=0.8\linewidth}}
    \centerline{(b) Rio in 11/10/2017.}\medskip
  \end{minipage}
  \hfill
  \begin{minipage}[b]{0.18\linewidth}
    \centering
    \centerline{\epsfig{figure=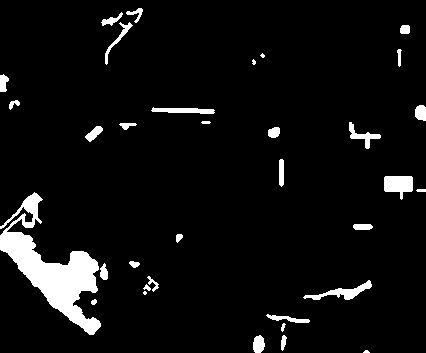,width=0.8\linewidth}}
    \centerline{(c) Ground truth.}\medskip
  \end{minipage}
  \hfill
  \begin{minipage}[b]{0.18\linewidth}
    \centering
    \centerline{\epsfig{figure=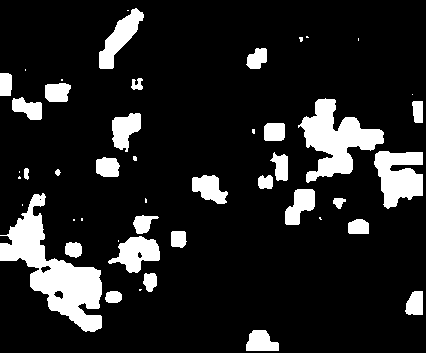,width=0.8\linewidth}}
    \centerline{(d) Early Fusion.}\medskip
  \end{minipage}
  \hfill
  \begin{minipage}[b]{0.18\linewidth}
    \centering
    \centerline{\epsfig{figure=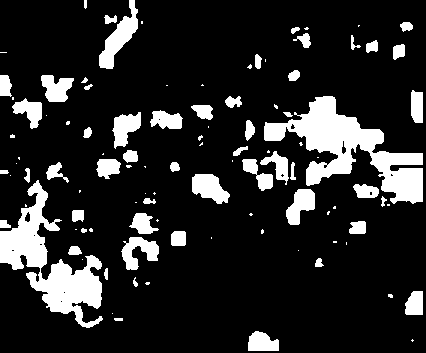,width=0.8\linewidth}}
    \centerline{(e) Siamese.}\medskip
  \end{minipage}
  
  \caption{Comparison between the results from the EF and Siamese architectures using 3 colour channels on the "Rio" test image.}
  \label{fig:arch_comp}
\end{figure*}

\begin{figure*}[ht]

  \begin{minipage}[b]{0.13\linewidth}
    \centering
    \centerline{\epsfig{figure=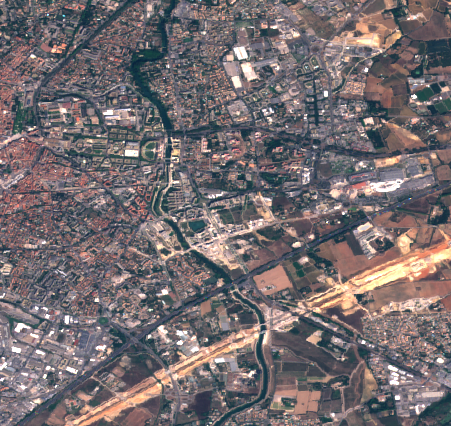,width=0.8\linewidth}}
    \centerline{(a) 12/08/2015.}\medskip
  \end{minipage}
  \hfill
  \begin{minipage}[b]{0.13\linewidth}
    \centering
    \centerline{\epsfig{figure=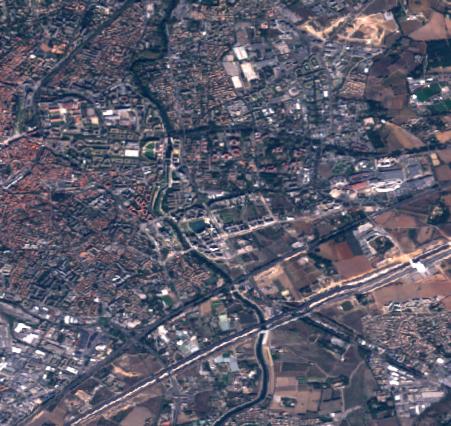,width=0.8\linewidth}}
    \centerline{(b) 30/10/2017.}\medskip
  \end{minipage}
  \hfill
  \begin{minipage}[b]{0.13\linewidth}
    \centering
    \centerline{\epsfig{figure=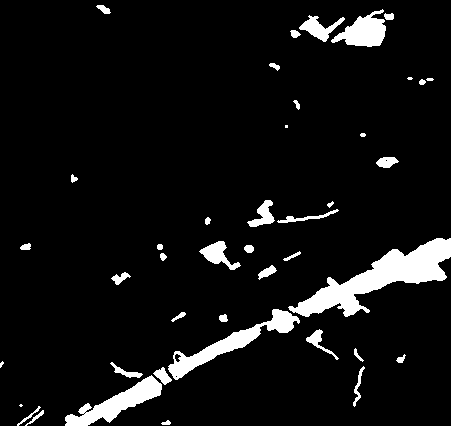,width=0.8\linewidth}}
    \centerline{(c) Ground truth.}\medskip
  \end{minipage}
  \hfill
  \begin{minipage}[b]{0.13\linewidth}
    \centering
    \centerline{\epsfig{figure=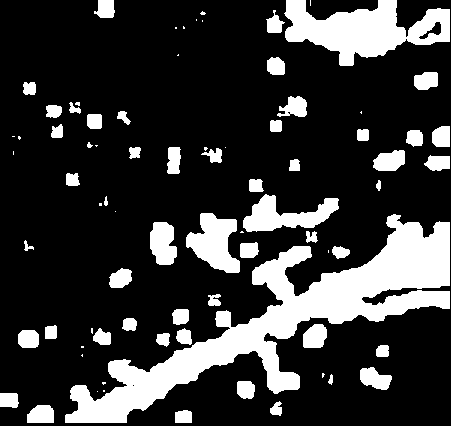,width=0.8\linewidth}}
    \centerline{(d) 3 channels.}\medskip
  \end{minipage}
  \hfill
  \begin{minipage}[b]{0.13\linewidth}
    \centering
    \centerline{\epsfig{figure=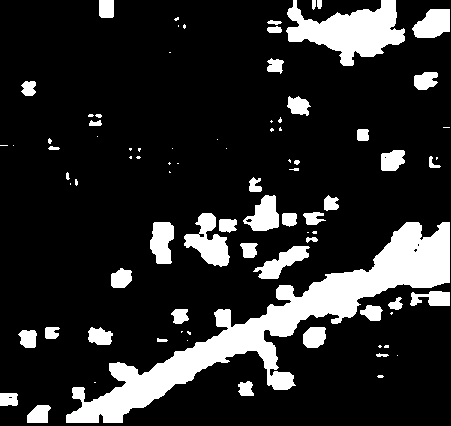,width=0.8\linewidth}}
    \centerline{(e) 4 channels.}\medskip
  \end{minipage}
  \hfill
  \begin{minipage}[b]{0.13\linewidth}
    \centering
    \centerline{\epsfig{figure=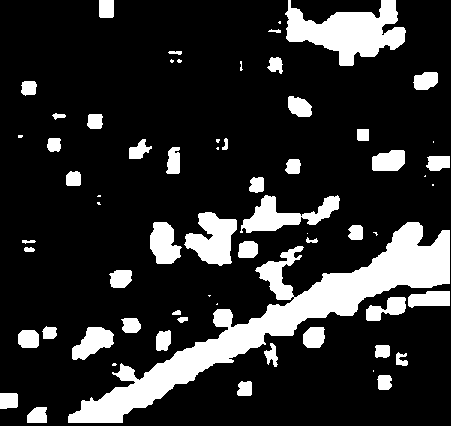,width=0.8\linewidth}}
    \centerline{(f) 10 channels.}\medskip
  \end{minipage}
  \hfill
  \begin{minipage}[b]{0.13\linewidth}
    \centering
    \centerline{\epsfig{figure=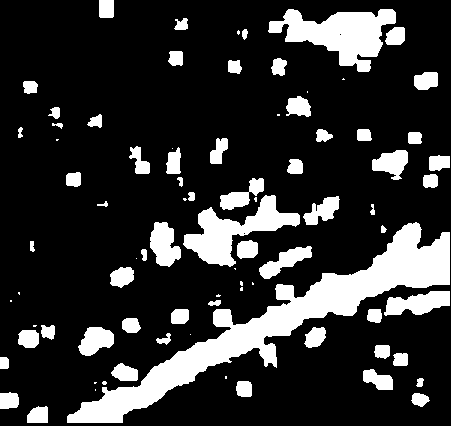,width=0.8\linewidth}}
    \centerline{(g) 13 channels.}\medskip
  \end{minipage}
  
  \caption{Comparison of results of the EF network on the "Montpellier" test image using 3, 4, 10 and 13 channels as input.}
  \label{fig:ch_comp}
\end{figure*}

The objective of the OSCD dataset is to provide an open and standardized way of comparing the efficacy of different change detection algorithms that are proposed by the scientific community, available to anyone who is interested in tackling the change detection problem. The dataset is focused on urban areas, labelling as Change only urban growth and changes and ignoring natural changes (e.g. vegetation growth or sea tides). Examples of image pairs and the associated change maps can be seen in Figs.~\ref{fig:arch_comp} and \ref{fig:ch_comp}. The dataset provides a comparison standard for single band, colour or multispectral change detection algorithms that are proposed. Since it contains pixel-wise ground truth change labels for each location on each image pair, the dataset also allows for more elaborate supervised learning methods to be applied to the problem of change detection.

The OSCD dataset was built using images from the Sentinel-2 satellites. The satellite captures images of various resolutions between 10m and 60m in 13 bands between ultraviolet and short wavelength infrared. Twenty-four regions of approximately 600x600 pixels at 10m resolution with various levels of urbanization where urban changes were visible were chosen worldwide. The images of all bands were cropped according to the chosen geographical coordinates, resulting in 26 images for each region, i.e. 13 bands for each of the images in the image pair.
These images were downloaded and cropped using the Medusa toolbox\footnote{\url{https://github.com/aboulch/medusa_tb}}.

The high variability of the raw data that is available from Sentinel-2 does not allow a completely scripted generation of image patches. The downloaded images frequently contain large sections of completely black pixels, and the correct images must be selected manually. Furthermore, for the generation of this dataset, it was desired to obtain images with no or very few clouds present in the image. While the \textit{sentinelsat} API allows some control over the amount of clouds present in the images, this also requires manual verification of each of the downloaded images to ensure the presence of clouds in the downloaded image is not too large.

The pixel-wise ground truth labels were manually generated by comparing the true colour images for each pair. To improve the accuracy of results, the GEFolki toolbox~\cite{brigot2016adaptation}\footnote{\url{https://github.com/aplyer/gefolki}} was used to register the images with more precision than the registration that is done by the Sentinel system itself. In all cases the older image in the pair was used as reference, and the newer one was transformed to perfectly align with it.

\subsection{Challenges and limitations}
\label{ssec:ds-challenges}

While the dataset is a very valuable tool for methodically comparing different change detection algorithms and for applying supervised learning techniques to this problem, it is important to understand the limitations of this dataset. First and foremost, the images generated by the Sentinel-2 satellite are of a relatively low resolution. This resolution allows the detection of the appearance of large buildings between the images in the image pair. Smaller changes such as the appearance of small buildings, the extension of existing buildings or the addition of lanes to an existing road, for example, may not be obvious in the images. For this reason, even the change maps that are generated manually by different analysts may differ.

One approach which was explored was using OpenStreetMap data from different dates to generate the change maps in an automated manner. OpenStreetMap provides open map data, and by comparing the maps for the dates of the images in the pairs it is, in theory, possible to identify what changes occurred in the area. This approach proved unsuccessful for a few reasons. First, most of the changes  in the maps between the two dates were actually due to things being added to the map, but which had not been built in the period between the dates when the images were taken. Second, it is not possible to have much precision when it comes to the dates of the older maps, where in many cases only one map was available for each year before 2017.

Finally, the Sentinel-2 satellite was launched in 2015, and therefore the data that is available is not able to go further in the past than June of 2015. This means that the changes contained in the dataset are of a temporal distance of at most two and a half years approximately, and less than that in many cases. This also means that the images contain many times more pixels labelled as no change than labelled as change.

\section{Change Detection Methods}
\label{sec:cd}

The aim of the methods presented in this section is to apply supervised deep learning methods to the change detection problem, training them only on the dataset presented in Section~\ref{sec:dataset}. Unlike previous methods which only use CNNs to build difference images which are later thresholded, our networks are trained end-to-end to classify a patch between two classes: change and no change. The patches are of size 15x15 pixels, and the networks attempt to classify the label of the central pixel based on its neighbourhood's values. The networks should ideally be able to learn to differentiate between artificialization changes and natural changes, given that only artificialization changes are labelled as changes on the dataset. This goes further than computing a simple difference between the images, as it involves a semantic interpretation of the changes, and is therefore a harder problem. 

\subsection{Architectures}
\label{ssec:cd-esc}

Two CNN architectures are compared for this work. These architectures are inspired by the work of Zagoruyko et al.~\cite{zagoruyko2015learning}, where similar architectures were used to compare image patches, although not with the purpose of detecting changes.
These networks take as input two 15x15xC patches, where C is the number of colour channels, which will be further discussed in Section \ref{sec:results}. The output of the networks for each pair of patches is a pair of values which are an estimation of the probability of that patch belonging to each class. By choosing the maximum of these two values we are able to predict if a change has occurred in the central pixel of the patch. Furthermore, we are able to threshold the change probability at values other than 0.5 to further control the results, in case false positives or false negatives are more or less costly in a given application.

The first proposed architecture, named Early Fusion (EF), consists of concatenating the two image pairs as the first step of the network. The input of the network can then be seen as a single patch of 15x15x2C, which is then processed by a series of seven convolutional layers and two fully connected layers, where the last layer is a softmax layer with two outputs associated with the classes of change and no change.

The second approach is a Siamese (Siam) network. The idea is to process each of the patches in parallel by two branches of four convolutional layers with shared weights, concatenating the outputs and using two fully connected layers to obtain two output values as before.

\subsection{Full image change maps}
\label{ssec:cd-cms}

Once the networks have been trained, full image change maps can be generated by classifying patches of the test images individually. To speed up this process, instead of taking a patch centred in each pixel of the image, a larger stride was used for extracting the patches and a voting system was implemented to predict the labels of all pixels in the images. Each classified patch votes on the label of all the pixels it covered based on the outputs of the network and with weight following a 2D Gaussian distribution centred on the central pixel of the patch, i.e. the closer a pixel is to the central pixel of a patch, the more that patch's vote will count when classifying that pixel.

\section{Results}
\label{sec:results}

The images of the OSCD dataset were split into two groups: fourteen images for training and ten for testing. Since the dataset is not very large, the training data was augmented by using all possible flips and rotations in steps of 90 degrees of the patches.

To deal with the problem of different resolutions in different channels, the channels with resolutions lower than 10m were upsampled to the resolution of 10m so that all channels could be concatenated with aligned pixels.

As was mentioned in Section~\ref{sec:dataset}, the number of pixels marked as having no changes is much larger than the number of pixels labelled as change. Thus, at training, we apply a higher penalization weight for the change class, the weights being inversely proportional to the number of examples of each class.

To evaluate the influence of the number of input channels in the classification, four cases were compared: color image (RGB, 3 channels), layers with resolution of 10m (RGB + infrared, 4 channels), layers with resolution up to 20m (10 channels), layers with resolution up to 60m (13 channels).

Table \ref{tab:results} explores eight variants of CNNs and shows their superiority to the difference image methods presented in~\cite{le2013urban}.
The first thing to observe in the results is that EF networks tend to perform better than their correspondent Siamese networks. It is also important to note that the addition of colour channels generally leads to an improvement in classification performance, but this does not happen linearly and each architecture reacts differently.

Figures \ref{fig:arch_comp} and \ref{fig:ch_comp} contain examples of change maps generated by some of the trained CNNs. They allow the comparison of results when using different architectures (Fig.~\ref{fig:arch_comp}) and different numbers of input channels (Fig.~\ref{fig:ch_comp}). These images indicate that the networks have succeeded in detecting urban changes in the images.


\begin{table}
\begin{tabular}{cc|c|c|c}

Data & Network & Acc. & Change acc. & No change acc.\\
\hline \hline
 3 ch. & Siam.      & 84.13 & 78.57 & 84.43 \\
       & EF         & 83.63 & 82.14 & 83.71 \\  \hline
 4 ch. & Siam.      & 75.20 & 74.71 & 75.23 \\
       & EF         &  \cellcolor{ForestGreen!35}\textbf{89.66} & 80.30 & \cellcolor{ForestGreen!35}\textbf{90.16} \\  \hline
10 ch. & Siam.      & 86.21 & \cellcolor{GreenYellow!35}83.04 & 86.38 \\
       & EF         & \cellcolor{YellowGreen!35}89.15 & 82.75 & \cellcolor{YellowGreen!35}89.50 \\  \hline
13 ch. & Siam.      & 85.37 & \cellcolor{ForestGreen!35}\textbf{85.63} & 85.35 \\
       & EF         & \cellcolor{GreenYellow!35}88.15 & \cellcolor{YellowGreen!35}84.69 & \cellcolor{GreenYellow!35}88.33 \\  \hline \hline
 \multicolumn{2}{c|}{Img. diff.} & 76.12 & 63.42 & 76.82 \\ 
    \multicolumn{2}{c|}{Log-ratio} & 76.93 & 59.68 & 77.87 \\ 
    \multicolumn{2}{c|}{GLRT}    & 76.25 & 60.48 & 77.11 \\  \hline
\end{tabular}
\caption{Evaluation metrics for each of the test cases. Accuracy here is the number of true positives for a given class divided by the total number of elements of that class on the test dataset.}
\label{tab:results}
\end{table}

\section{Conclusion}
\label{sec:conclusion}

In this paper, we presented the Onera Satellite Change Detection dataset, the first one for urban change detection made of Sentinel-2 images and openly available. We also presented two CNN approaches for detecting changes on image pairs of this dataset. These networks were trained in a fully supervised manner with no use of other image data and obtain excellent test performances.


Perspectives for this work include enlarging the dataset both in number of cities and imaging modalities, e.g. enriching it with Sentinel-1 satellites images.
It is also a logical next step to experiment with fully convolutional networks to automatically generate labels for all pixels in the image, reducing the patch effect of the presented methods. It would also be of interest to extend this work to deal with semantic labelling of changes, which would be of even more help in interpreting the image pairs. Finally, it would be useful to be able to handle image sequences for change detection instead of image pairs.

\bibliographystyle{IEEEbib}
\bibliography{strings,refs}

\end{document}